\documentclass[a4paper,fleqn]{cas-dc}

\usepackage{lipsum}
\usepackage{lineno}
\usepackage{multirow}
\usepackage{graphicx}
\usepackage{grffile}
\usepackage{epstopdf}

\usepackage{upgreek}
\usepackage{bm}

\usepackage{ragged2e}
\usepackage{graphicx}
\renewcommand{\raggedright}{\leftskip=0pt \rightskip=0pt plus 0cm}

\usepackage{amsmath,amsthm}
\usepackage{setspace}

\usepackage[numbers]{natbib}
\usepackage{amsfonts}
\usepackage{amsmath}
\xspaceaddexceptions{]}
\def\tsc#1{\csdef{#1}{\textsc{\lowercase{#1}}\xspace}}
\tsc{WGM}
\tsc{QE}
\tsc{EP}
\tsc{PMS}
\tsc{BEC}
\tsc{DE}

\begin{document}
\let\WriteBookmarks\relax
\def\floatpagepagefraction{1}
\def\textpagefraction{.001}
\shorttitle{Universal Approximation with Quadratic Deep Networks}
\shortauthors{Fan et al.}

\title [mode = title]{Universal Approximation with Quadratic Deep Networks}                      

\author[1]{Fenglei Fan}

\credit{Rensselaer Polytechnic Institute}

\address[1]{Department of Biomedical Engineering, Rensselaer
Polytechnic Institute, Troy, NY, 12180 USA }

\author[2]{Jinjun Xiong}
\author[1]{Ge Wang}

\cormark[1]
\ead{wangg6@rpi.edu}

\cortext[cor1]{Corresponding author}

\address[2]{IBM Thomas J. Watson Research Center, Yorktown Heights, NY, 10598, USA}

\begin{abstract}[S U M M A R Y]
Recently, deep learning has achieved huge successes in many important applications. In our previous studies, we proposed quadratic/second-order neurons and deep quadratic neural networks. In a quadratic neuron, the inner product of a vector of data and the corresponding weights in a conventional neuron is replaced with a quadratic function. The resultant quadratic neuron enjoys an enhanced expressive capability over the conventional neuron. However, how quadratic neurons improve the expressing capability of a deep quadratic network has not been studied up to now, preferably in relation to that of a conventional neural network. Regarding this, we ask four basic questions in this paper:
(1) for the one-hidden-layer network structure, is there any function that a quadratic network can approximate much more efficiently than a conventional network?
(2) for the same multi-layer network structure, is there any function that can be expressed by a quadratic network but cannot be expressed with conventional neurons in the same structure?
(3) Does a quadratic network give a new insight into universal approximation? 
(4) To approximate the same class of functions with the same error bound, is a quantized quadratic network able to enjoy a lower number of weights than a quantized conventional network? 
Our main contributions are the four interconnected theorems shedding light upon these four questions and demonstrating the merits of a quadratic network in terms of expressive efficiency, unique capability, compact architecture and computational capacity respectively.
\end{abstract}

\maketitle

\section{Introduction}

Over recent years, deep learning has become the mainstream approach for machine learning. Since AlexNet [1], increasingly more advanced neural networks [2-6] are being proposed, such as GoogleNet, ResNet, DenseNet, GAN and their variants, to enable practical performance comparable to or beyond what human delivers in computer vision [7],
speech recognition [8], language processing [9] game playing [10], medical imaging [11-13], and so on.
A heuristic understanding of why these deep learning models are so successful is that these models represent knowledge in hierarchy and facilitate high-dimensional non-linear functional fitting. It seems that deeper structures are correlated with greater capacities to approximate more complicated functions. \\

\indent The representation power of neural networks has been rigorously studied since the eighties. The first result is that a network with a single hidden layer can approximate a continuous function at any accuracy given an infinite number of neurons [14]. That is, the  network can be extremely wide. With the emergence of deep neural networks, studies have been performed on theoretical benefits of these deep models over shallow ones [15-24]. One way [15, 19] is to construct a special kind of functions that are easy to be approximated by deep networks but hard by shallow networks. It has been reported in [16] that a fully-connected network with ReLU activation can approximate any Lebesgue integrable function in the $L_1$-norm sense, provided a sufficient depth and at most $d+4$ neurons in each layer, where $d$ is the number of inputs. Through a similar analysis, it was reported in [22] that ResNet with one single neuron per layer is a universal approximator. Moreover, it was demonstrated in [15] that a special class of functions is hard to be approximated by a conventional network with a single hidden layer unless an exponential number of neurons are used. Bianchini et al. [23] utilized a topological measure to characterize the complexity of functions that are realized by neural networks, and proved that deep networks can represent functions of much higher complexity than the shallow counterparts. Kurkova et al. [24] showed that unless with sufficiently many network units (more than any polynomial of the logarithm of the size
of the domain), a good approximation cannot be achieved with a shallow perceptron network for almost any uniformly chosen function on a specially constructed domain. \\

\indent In our previous studies [25-28], we proposed quadratic neurons and deep quadratic networks. In a quadratic neuron, the inner product of an input vector and the corresponding weights in a conventional neuron is replaced with a quadratic function. The resultant quadratic neuron enjoys an enhanced expressive capability over the conventional neuron (Here conventional neurons refer to neurons that perform activation of an inner-product, and conventional networks refer to networks completely comprising conventional neurons). Actually, a quadratic neuron can be viewed as a fuzzy logic gate, and a deep quadratic network is nothing but a deep fuzzy logic system [27]. Furthermore, we successfully designed a quadratic autoencoder for a real-world low-dose CT problem  [28].  Note that high-order neurons [29-30] were taken into account in the early stage of artificial intelligence, but they are not connected to deep networks and suffer from a combinatorial explosion with the number of parameters due to high order terms. In contrast, our quadratic neuron uses a limited number of parameters (tripled that of a conventional neuron) and performs a cost-effective high-order operation in the context of deep learning. For quadratic deep networks, we already developed a general backpropagation algorithm [26], enabling the network training process.  \\

\indent However, how quadratic neurons improve the expressing capability of a deep quadratic network has not been theoretically studied up to now, preferably in relation to that of a conventional neural network. In this paper, we ask four basic questions regarding the expressive capability of a quadratic network:
(1) for the one-hidden-layer network structure, is there any function that a quadratic network can approximate much more efficiently than a conventional network?
(2) for the same multi-layer network structure, is there any function that can be expressed by a quadratic network but cannot be expressed with conventional neurons in the same structure?
(3) Does a quadratic network give a new insight and a new method for universal approximation? 
(4) To approximate the same class of functions with the same error bound, is a quadratic network able to enjoy a lower number of weights than a conventional network? 
If the answers to these questions are favorable, quadratic networks should be significantly more powerful in many machine learning tasks.\\

\indent In this paper, our contributions are to present four theorems addressing the above questions respectively and positively, thereby establishing the intrinsic advantages of quadratic networks over conventional networks. More specifically, these theorems characterize the merits of a quadratic network in terms of expressive efficiency, unique representation, compact architecture and computational capacity. We answer the first question with the first theorem, given the network with only one hidden layer and admissible activation function, there exists a function that a quadratic network can approximate it with a polynomial number of neurons but a conventional network can only do the same level approximation with an exponential number of neurons.Regarding the second question and the second theorem below, with the ReLU activation function, any continuous radial function can be approximated by a quadratic network in a structure of no more than four neurons in each layer but the function cannot be approximated by a conventional network of the same structure [16], while with the third theorem below we provide a new insight and a new method for a universal presentation from the perspective of the Algebraic Fundamental Theorem. Without introducing complex numbers, a univariate polynomial of degree $n$ can be uniquely factorized into a product of quadratic terms instead of first-order terms. Since a ReLU quadratic network can represent any univariate polynomial in a \textit{unique and global} manner, and by the Weierstrass theorem and the Kolmogorov theorem that multivariate functions can be represented through summation and composition of univariate functions, we can approximate any multivariate function with a well-structured ReLU quadratic neural network, justifying the universal approximation power of the quadratic network. To our best knowledge, our quadratic network is the first-of-the-kind global universal approximator. Finally, with the fourth theorem below, we attempt to address the theoretical foundation for quantization of a neural network. To approximate the same class of functions with the same error bound, a quantized quadratic network demands a much lower number of weights than a quantized conventional network. The saving is almost in the order of $\mathcal{O}\left(\lambda\left(log^{\frac{1}{\lambda-1}+1}(\frac{1}{\epsilon})\right)(\frac{1}{\epsilon})^{\frac{d}{n}}\right)$, $\lambda \geq 2$, $d,n \geq 1$. Compared to the previous studies that theoretically explored the properties of quantized networks, our result adds unique insights into this aspect. \\

There are prior papers related to but different from our contributions [40-45]. Motivated by a need for more powerful activation, Livni et al. [40] and Krotov et al. [41] proposed to use the quadratic activation: $\sigma(z)=z^2$ and rectified polynomials in the neuron respectively. Despite somewhat misleading in their name, networks with quadratic activation or rectified polynomials and our proposed networks that consist of quadratic neurons have fundamental differences. At the cellular level, a neuron with quadratic activation is still characterized with a linear decision boundary, while our quadratic neuron allows a quadratic decision boundary. In [40], the authors demonstrated that networks with quadratic activation are as expressive as networks with threshold activation, and constant-depth networks with quadratic activation can learn in polynomial time. In contrast, our work goes further showing that the expressibility of the quadratic network is superior to that of the conventional networks; for example, a single quadratic neuron can implement the XOR gate, and a quadratic network of finite width and depth can represent a finite-order polynomial up to any desirable accuracy. In [42], Du et al. showed that over-parameterization and weight decay are instrumental to the optimization aided by quadratic activation. [43] reported how a neural network can provably learn a low-degree polynomial with gradient descent search from scratch, with an emphasis on the effectiveness of the gradient descent method. [44] presents that $O(log \frac{p}{\epsilon})$ layers of binary units and  $O(p log \frac{p}{\epsilon})$ 
ReLU units can approximate $f(x)=\sum_{i=0}^{p} a_ix^i$ with closeness of $\epsilon$. In contrast, our Theorem 3 is based on the Algebraic Fundamental Theorem to provide an exact representation of any finite-order polynomial. [45] is on factorization machine (FM) dedicated to combine high order features, clearly different from the polynomial factorization we propose to perform using a quadratic network.

\section{Preliminaries}

\indent \textbf{Quadratic/Second-order Neuron}: The $n$-input function of a quadratic/second-order neuron before being nonlinearly processed is expressed as:
\begin{equation}
\begin{aligned}
h(\textbf{x})&=(\sum_{i=1}^{n} w_{ir}x_i +b_r)(\sum_{i=1}^{n} w_{ig}x_i +b_g) + \sum_{i=1}^{n} w_{ib}x_{i}^2+c \\
&=(\textbf{w}_{r}\textbf{x}^T+b_{r})(\textbf{w}_{g}\textbf{x}^T+b_{g})+\textbf{w}_{b}(\textbf{x}^2)^T+c,
\end{aligned}
\end{equation}
where $\bf x$ denotes the input vector, $\bf w_r,\bf w_g, \bf w_b$ are vectors of the same dimension with $\bf x$ and $b_r, b_g, c$ are adjustable biases.
Our quadratic function definition only utilizes $3n$ parameters, which is more compact than the general second-order representation requiring $\frac{n(n+1)}{2}$ parameters. While our quadratic neuron design is unique, other papers on quadratic neurons are also in the later literature; for example, 
[31] proposed a type of neurons with paraboloid decision boundaries. It is underlined that the emphasis of our work is not only on quadratic neurons individually but also deep quadratic networks in general.\\
\indent \textbf{One-hidden-layer Networks:} The generic function represented by a one-hidden-layer conventional network is as follows:
\begin{equation}
\begin{aligned}
f_1(\textbf{x}) 
=\sum_{l=1}^{k} t^{l}\sigma_{l} (\textbf{w}^l\textbf{x}+b^{l}),
\end{aligned}
\end{equation}
where $l$ refers to the $l^{th}$ layer. In contrast, the generic function represented by a one-hidden-layer quadratic network is:
\begin{equation}
\begin{aligned}
f_2(\textbf{x})=\sum_{l=1}^{k} t^{l}\sigma_{l}[(\textbf{w}_{r}^{l}\textbf{x}^T+b_{r}^{l})(\textbf{w}_{g}^{l}\textbf{x}^T+b_{g}^{l})+\textbf{w}_{b}^{l}(\textbf{x}^2)^T+c^l].
\end{aligned}
\end{equation}
In our Theorem 1 below, we will compare the representation capability of a quadratic network and that of a conventional network assuming that both networks have the same one-hidden-layer structure. \\
\indent \textbf{$L$-Lipschitz Function}: A $L$-Lipschitz function $f$ from $\mathbb{R}^n$ to $\mathbb{R}$ is defined by the following property:
\begin{equation*}
|f(\textbf{x})-f(\textbf{y})|\leq L||\textbf{x}-\textbf{y}||
\end{equation*}
\indent \textbf{Radial Function}: A radial function only depends on the norm of its input vector, generically denoted as $f(||\textbf{x}||)$. The functions mentioned in  Theorems 1 and 2 are all radial functions. By its nature, the quadratic neuron is well suited for modeling of a radial function.\\
\indent \textbf{Euclidean Unit-volume Ball}: In a $d$-dimensional space, let $R_d$ be the radius of a Euclidean ball $B_d$ such that $R_{d}B_{d}$ has the unit volume. Euclidean ball is used to define density function $\mu$ later. \\
\indent \textbf{Bernstein Polynomial:} $l_{n,m}=C_{n}^{m}x^{m}(1-x)^{n-m}, 0 \leq m \leq n$. The $n$-th Bernstein polynomial of $f(x)$ in $ (0,1) $ is defined as
\begin{equation*}
B_{n}(x)=\sum_{m=0}^{n}f(\frac{m}{n})l_{n,m}(x)
\end{equation*}
Previously, the Bernstein Polynomials are used to prove the Weierstrass theorem. \\
\indent \textbf{Admissible function:} An activation function $\sigma: \mathbb{R} \to \mathbb{R}$ is called admissible if:\\
\indent 1) $\sigma$ is measurable \\
\indent 2) there are $K, \kappa>0$, $|\sigma(x)| \leq K(1+|x|^{\kappa})$ for all $x\in \mathbb{R}$\\
\indent 3) given any $L-$Lipschitz function $f:\mathbb{R} \to \mathbb{R}$ which is constant outside a bounded interval $[-R,R]$ and any $\delta >0$, there are scalars $a\in \mathbb{R}$ and $(\alpha_i,\beta_i, \gamma_i)_{i=1,2,...,w}$ such that 
\begin{equation*}
|f(x)-[a+\sum_{i=1}^{w}\alpha_i \sigma(\beta_ix-\gamma_i)]| \leq \delta,
\end{equation*}
where $w\leq c_{\sigma}\frac{RL}{\delta}$.\\
\indent The standard activation functions such as ReLU and sigmoid satisfy the above three properties. Detailed proofs are out of the scope of this paper, please refer to [15,45]. \\
\indent \textbf{Function space $\mathcal{F}_{n}^{d}$}: The Sobolev space $\mathcal{W}^{n,\infty}([0,1]^d)$ with $n=1, 2,...$ is defined on $[0,1]^d$, lying in $L^{\infty}$ together with their weak derivatives of up to order $n$.  The function space $\mathcal{F}_{n}^{d}$ is made of any function $f \in \mathcal{W}^{n,\infty}([0,1]^d)$ and 
\begin{equation*}
\max_{\textbf{n}:|\textbf{n}| \leq n} {ess\sup}_{\textbf{x} \in [0,1]^d} |D^{\textbf{n}}f(\textbf{x})| \leq 1,
\end{equation*}
where $|\textbf{n}|=\sum_{k=1}^{d} n_k$. 

\section{Four Theorems}
First, we present four theorems, and then give their proofs respectively. \\
\indent \textit{Theorem 1:} \textit{For an admissible activation function $\sigma(\cdot)$ and for some universal constants $c>0, C>0, C^{'}>0, c_\sigma>0$, there exist a probability measure $\mu$ and a radial function $\tilde{g}$: $\mathbb{R}^d \to\mathbb{R}$, where $d>C$, that is bounded on [-2,2] and supported on $||\textbf{x}|| \leq C^{'}\sqrt{d}$ satisfying:\\
\indent 1. $\tilde{g}$ can be approximated by a single-hidden-layer quadratic network with $C^{'}c_\sigma d^{3.75}$ neurons, which is denoted as $f_2$. \\
\indent 2. For every function $f_1$ expressed by a single-hidden-layer conventional network with at most $ce^{cd}$ neurons, we have: }
\begin{equation*}
E_{x\sim \mu} (f_1(\textbf{x})-f_2(\textbf{x}))^2 \geq \delta
\end{equation*}
for some positive constant $\delta$. \\
\indent \textit{Theorem 2: For any compactly supported, radial, continuous function $f$: $\mathbb{R}^n \to \mathbb{R}$ and any $\epsilon>0$, there exists a function that can be implemented by a ReLU-activated quadratic network with at most four neurons in each layer, such that:}
\begin{equation*}
\sup_{x \in \mathbb{R}^d} |f(x)-g(x)| dx \leq \epsilon.
\end{equation*}

\indent \textit{Theorem 3:} \textit{With ReLU as activation function, the quadratic network is a global universal approximator.} \\
\indent \textit{Theorem 4:} \textit{For any $f \in \mathcal{F}_{d}^{n}$ and any $\epsilon \in (0,1)$, there is a ReLU quadratic network with $\lambda$ distinct weights that can approximate $f$ with $\epsilon$, satisfying (i) the depth is $\mathcal{O}\left(log(log(\frac{1}{\epsilon}))+log(\frac{1}{\epsilon})\right)$ (ii) the number of weights is \\$ \mathcal{O}\left(log(log(\frac{1}{\epsilon}))(\frac{1}{\epsilon})^{\frac{d}{n}}+ log(\frac{1}{\epsilon})(\frac{1}{\epsilon})^{\frac{d}{n}}\right)$.} \\


\textbf{\large{3.1 Theorem 1}}\\
\indent \textbf{Key Idea for Proving Theorem 1:} The form of functions represented by a single-hidden-layer conventional network is $f_1(\textbf{x})=\sum_{l=1}^{k} t^{l}\sigma_{l} (\textbf{w}^l\textbf{x}+b^{l})$. It is observed that the distribution of the Fourier transform of $f_1(\textbf{x})$ is supported on a finite collections of lines. The support covered by the finite lines are sparse in the Fourier space, especially for a high dimension and high frequency regions, unless an exponential number of lines are involved. Thus, a possible target function to be constructed should have major components at high-frequencies. A suitable candidate has been constructed in [15]:
\begin{equation*}
\tilde{g}(\textbf{x})=\sum_{i=1}^{N} \epsilon_{i}g_{i}(||\textbf{x}||),
\end{equation*}
where $\epsilon_i \in \{-1,1\}$, $N$ is a polynomial function of $d$, $g_i(||\textbf{x}||)=\textbf{1}\{||\textbf{x}||\in \Delta_i\}$ are radial indicator functions over disconnected intervals. According to the definition, $\tilde{g}$ is well bounded.  Although the constructed $\tilde{g}$ is hard to approximate by a conventional network, it is easy to approximate by a quadratic network, because $\tilde{g}$ is a radial function. Consequently, it is  feasible for a single-hidden-layer quadratic network to approximate the radial function with a polynomial number of neurons. Note that $\tilde{g}(\textbf{x})$ is discontinuous, hence it cannot be perfectly expressed by a neural network with continuous activation functions. Therefore, let us use a probability measure $\mu$ for network quality assessment.  With $\mu$, the distance between $\tilde{g}$ and $f$ represented by a network is characterized as $ E_{x\sim \mu} (f(\textbf{x})-\tilde{g}(\textbf{x}))^2$. In particular, $\mu =\phi^2$, where $\phi$ is the inverse Fourier transform of the indicator function defined on the Euclidean ball: $\textbf{1}\{x \in B\}$. Therefore, $ E_{x\sim \mu} (f(\textbf{x})-\tilde{g}(\textbf{x}))^2 = ||\widehat{f\phi}-\widehat{\tilde{g}\phi}||_{L_2}$, where $\widehat{f\phi}$, $\widehat{\tilde{g}\phi}$ are the frequency forms of $f\phi$ and $\tilde{g}\phi$. The physical meaning of $E_{x\sim \mu} (f(\textbf{x})-\tilde{g}(\textbf{x}))^2$ is to measure the closeness of $\tilde{g}$ and $f$ in frequency domain within Euclidean ball $B$. \\ 

\indent \textbf{The contribution of our work:} In our Theorem 1, we demonstrate the utility of a quadratic network in approximating the constructed radial function $\tilde{g}$, which is straightforward but interesting and non-trivial. Furthermore, our result holds for all the admissible activation functions, while the proof in [15] is only for the ReLU activation function. Proposition 1 in [15] demonstrates that $\tilde{g}$ cannot be well approximated by a single-hidden-layer conventional network with a polynomial number of neurons. We put this proposition here as Lemma 1 for completeness.  We would like to emphasize that despite we put Lemma 1 and other lemmas on an equal footing, actually Lemma 1 is more important for our theorem. Again, our contribution is the proof that the one-hidden-layer quadratic network using an admissible activation function can approximate $\tilde{g}$ efficiently.\\

\indent \textbf{Lemma 1:} There are universal constants $c,C>0$ such that for every $d>C$, there is a probability measure $\mu$ on $\mathbb{R}^d$ such that for any $\alpha>C$ and $N \geq C\alpha^{1.5}d^2$, there exists a function $\tilde{g}(\textbf{x})=\sum_{i=1}^{N} \epsilon_{i}g_{i}(\textbf{x})$, where $\epsilon_i \in \{-1,1\}, i=1,2,...,N$, such that for any function of the form: $f_1(\textbf{x}) =\sum_{l=1}^{k} t^{l}\sigma_{l} (\textbf{w}^l\textbf{x}+b^{l})$ with $\sigma_{l}$ as admissible function and $k \leq ce^{cd}$, we have: \\ 
\begin{equation*}
||f_1-\tilde{g}||_{L_{2}(\mu)} \geq \delta/\alpha,
\end{equation*}
where $\delta > 0$ is a universal constant. \\

Universal constants means they are independent from the dimension $d$. To illustrate $\tilde{g}$ is expressible with a quadratic network, we know from Lemma 12 of [15] that a continuous Lipschitz function $g$ can approximate $\tilde{g}$, what is remained for us to do is to use a quadratic network with a polynomial number of neurons to approximate $g$.\\


\indent \textbf{Lemma 2:} Given an admissible activation function  $\sigma$, there is a constant $c_\sigma \geq$ 1 (depending on $\sigma$ and other parameters) such that 
for any L-Lipschitz function $g: \mathbb{R} \rightarrow \mathbb{R}$,  which is constant outside a bounded interval [r, R] ($r \geq 1$) and any $\delta$, there exist scalars $a$,  \{$\alpha_i, \beta_i, \gamma_i$ \}, $i=1, \dots, w$, $w\leq c_{\sigma}\frac{(R^2-r^2)L}{4\sqrt{r}\delta}$ with which we have:
\begin{equation*}
h(||\textbf{x}||^2)=a+\sum_{i=1}^{w} \alpha_i\sigma(\beta_i ||\textbf{x}||^2-\gamma_i)
\end{equation*} 
satisfies:
\begin{equation*}
\mathop{\sup}_{||x|| \in [r,R]} \ \  |g(||\textbf{x}||)-h(||\textbf{x}||^2)|<\delta.
\end{equation*}

\indent Proof: $g:\mathbb{R} \to \mathbb{R}$ is an L-Lipschitz function and supported on $[r,R]$, $r > 0$. Let us define $s(x)=g(\sqrt{x})$, which is of $\frac{L}{2\sqrt{r}}$-Lipschitz and supported on $[r^2, R^2]$. By the property of an admissible function, we construct a function $h(t)=a+\sum_{i=1}^{w} \alpha_i\sigma(\beta_i t-\gamma_i)$, which satisfies the following condition:
\begin{equation*}
\mathop{\sup}_{l \in [r^2,R^2]} \ \  |s(l)-h(l)|<\delta,
\end{equation*}
where $w=c_{\sigma}\frac{(R^2-r^2)L}{4\sqrt{r}\delta}$.
Next, we have: 

\begin{equation*}
\begin{split}
&\sup_{||\textbf{x}|| \in [r,R]} |g(||\textbf{x}||)-h(||\textbf{x}||^2)|  \\
=&\sup_{||\textbf{x}|| \in [r,R]} |s(||\textbf{x}||^2)-h(||\textbf{x}||^2)|  \\
=&\sup_{l \in [r^2,R^2]} |s(l)-h(l)| < \delta
\end{split}
\end{equation*}

\indent \textbf{Lemma 3:} \textit{There are a universal constant $C>0$ and $\delta \in (0,1)$, for $d \geq C$ and any choice of $\epsilon_i \in \{-1,1\}, i=1,2,...,N$, there exists a function $f_2$ expressed by a single-hidden-layer quadratic network of a width of at most $c_\sigma \frac{3N \alpha^{1.5} d^{1.75}}{4\delta}$ and with the range [-2,+2] such that}
\begin{equation*}
||f_2-\tilde{g}||_{L_{2}(\mu)} \leq \frac{\sqrt{3}}{\alpha d^{\frac{1}{4}}}+\delta.
\end{equation*}

\indent Proof: In  Lemma 2, we make the following substitutions: $R=2\alpha \sqrt{d}$, $r=\alpha \sqrt{d}$, $L=N$, $h=g$. Notably, $\alpha$ can be greater than 1, and $r=\alpha \sqrt{d}>1$ satisfying the condition of Lemma 2. Thus, $g(||\textbf{x}||)$ is expressible by a single-hidden-layer quadratic network with at most $c_\sigma \frac{3N \alpha^{1.5} d^{1.75}}{4\delta}$ neurons. Coupled with Lemma 12 of [15], Lemma 3 is immediately obtained. \\
\\
\indent \textbf{Proof of Theorem 1:} By the combination of Lemmas 1 and 3, the proof for Theorem 1 is straightforward. In Lemma 1, by choosing $\alpha=C$, $N=C\alpha^{1.5}d^{2}$, we have
\begin{equation*}
||f_1-\tilde{g}||_{L_{2}(\mu)} \geq \delta_1
\end{equation*}
Let $\delta \leq \frac{\delta_1}{2} - \frac{\sqrt{3}}{Cd^{\frac{1}{4}}}$, which demands $d$ is very large so that $\delta>0$, to approximate $\tilde{g}$ we need the number of quadratic neurons being at most
\begin{equation*}
c_\sigma \frac{3N \alpha^{1.5} d^{1.75}}{4\delta}=c_\sigma \frac{3C^{2.5} d^{3.75}}{4\delta} \leq C^{'}c_\sigma d^{3.75}
\end{equation*}
(where $C^{'}$ is a universal constant depending on the universal constants $C$,$\delta_1$)

such that 
\begin{equation*}
||f_2-\tilde{g}||_{L_{2}(\mu)} \leq \frac{\sqrt{3}}{Cd^{\frac{1}{4}}}+\delta \leq \frac{\delta_1}{2}.
\end{equation*} 
Therefore, we have $||f_1-f_2||_{L_{2}(\mu)} \geq \frac{\delta_1}{2}$. The proof is completed. \\

\indent \textbf{Classification Example:} The above exponential difference between the conventional and quadratic networks is due to the dimension $d$. To demonstrate this, we constructed an example for separation of two concentric rings. In this example, there are 60 instances in each of the two rings representing two classes. With only one quadratic  neuron in a single hidden layer, the rings were totally separated, while at least six conventional neurons are required to complete the same task, as shown in Figure 1.\\

\begin{figure}
\center{\includegraphics[height=1.7in,width=3.5in,scale=0.4] {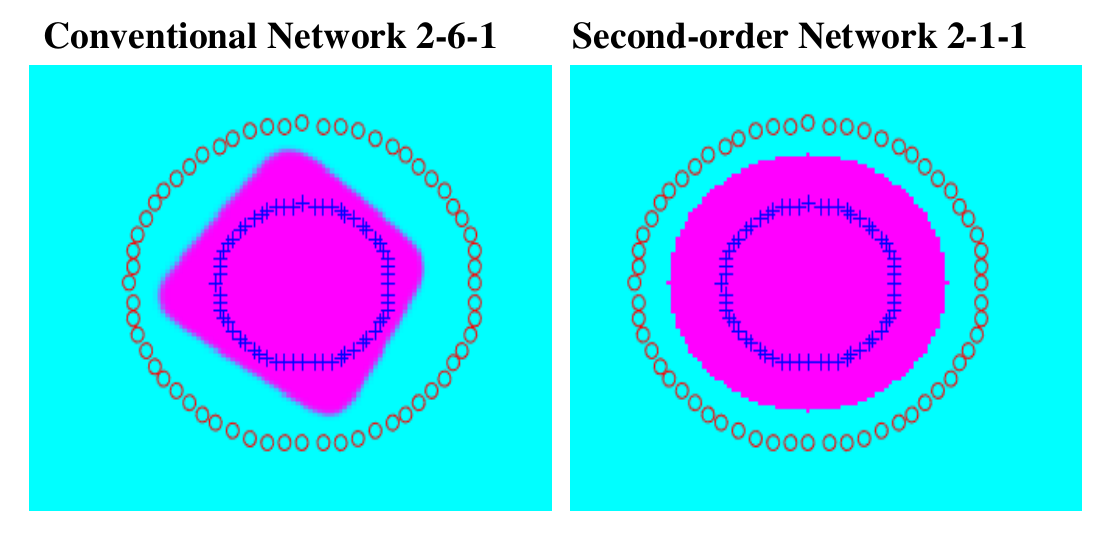}}
\caption{Classification of two concentric rings with conventional and quadratic networks (d=2). To succeed in the classification, a conventional network requires at least 6 neurons (Left), and a quadratic network only takes one neuron.}
	
\end{figure}

\textbf{\large{3.2 Theorem 2}}\\
\indent \textbf{Key Idea for Proving Theorem 2:} It was proved in [16] that an $n$-input function that is not constant along any direction cannot be well approximated by a conventional network with no more than $n-1$ neurons in each layer. However, for such a radially defined function, it is feasible for a quadratic network to approximate the function with width=4, which breaks the lower width bound $n+4$ given in [16]. The trick for the approximation by a deep conventional network is adapted for analysis of a deep quadratic network so that the function can be approximated in a "quadratic" way via composition layer by layer. To compute a radial function, we need to find the norm and then evaluate the function approximation by the norm. With a quadratic neuron, the norm can be easily found. Heuristically speaking, a quadratic network with no more than $n-1$ neurons in each layer could approximate a radial function very well, even if the function is not constant along any direction. Therefore our theorem positively shows that it is more natural and more effective to represent functional non-linearity with a quadratic network. Finally, since a quadratic network with width at most 4 can approximate any radial function and radial function family can approximate any continuous function [47], a quadratic network can serve as a width-bounded universal approximator. \\

\begin{figure}
  \centering
  \includegraphics[scale=0.35]{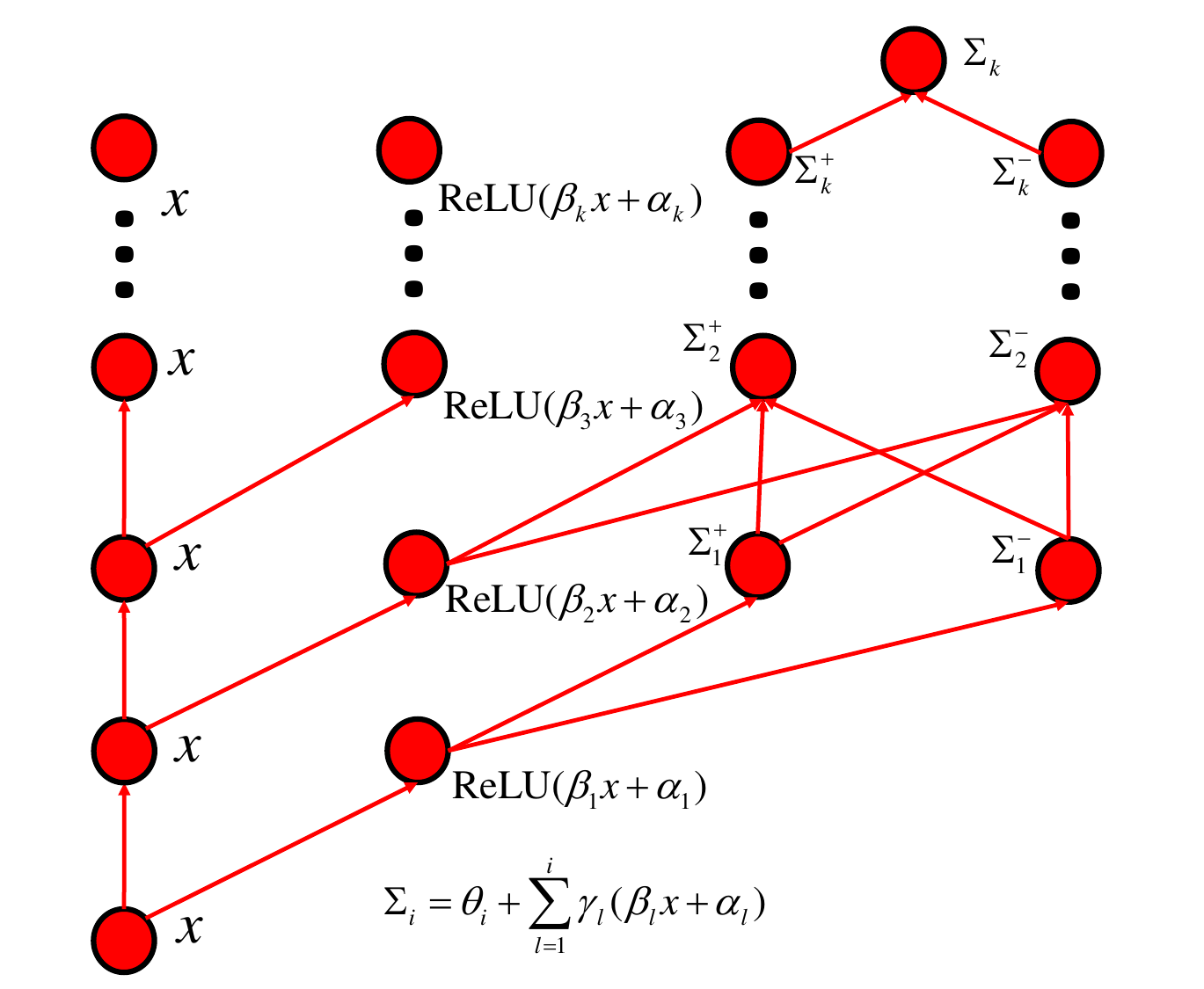}  
  \caption{Illustration of a conventional network structure that allows to implement any continuous univariate piecewise linear function}
\end{figure}

\textbf{Proof of Theorem 2:}  As per structure outlined in Figure 2, a conventional ReLU network with four neurons in each layer can implement any function of the form:
\begin{equation*}
[0, +\infty] \to \mathbb{R} : x \to \theta + \sum_{l=1}^{k} \gamma_l ReLU(\alpha_l x+ \beta_l)
\end{equation*},
which is actually any continuous piecewise linear function defined on non-negative real number domain.  Because function $f: \mathbb{R}^d \to \mathbb{R}$ is radial and continuous, without loss of generality, 
Let $h(||\textbf{x}||^2)=f(\textbf{x})$, while $h$ is function: $[0,+ \infty ]\to \mathbb{R}$. It implies that for any $\epsilon>0$, there will be a continuous piecewise linear function $g: [0,+ \infty ] \to \mathbb{R}$, and such that 
\begin{equation*}
\sup_{[0,+ \infty ]}|h(s)-g(s)|< \epsilon.
\end{equation*}

$g$ can be implemented by a conventional ReLU network as shown in Figure 2. Therefore, the function $\textbf{x} \to g(||\textbf{x}||_2^2): \mathbb{R}^d \to \mathbb{R}$ can be implemented with a quadratic ReLU network of the same structure by calculating norm of input. Then we will obtain:
\begin{equation*}
\begin{split}
&\sup_{x \to \mathbb{R}^d} |f(\textbf{x})-g(||\textbf{x}||_2^2)|  \\
&=\sup_{x \to \mathbb{R}^d} |h(||\textbf{x}||_2^2)-g(||\textbf{x}||_2^2)| \\
&=\sup_{s \in [0,+ \infty ]} |h(s)-g(s)|  \\
& < \epsilon
\end{split}
\end{equation*}

\textbf{Lemma 4:} [47] Let $K: \mathbb{R}^r \to \mathbb{R}$ be an integrable bounded function such that $K$ is continuous almost everywhere and $\int_{\mathbb{R}^r} K(x)\,dx\ \neq 0$. Then the functional family $S_K: \sum_{i=1}^{M}w_i K(\frac{||x-z_i||}{\sigma})$,($M$ is a positive integer, $\sigma >0$ and $w_i \in \mathbb{R}$) is dense in $L^p(\mathbb{R}^r)$. \\
\indent Proof: Please see [Theorem 1, 47].\\

\textbf{Corollary 2:} With ReLU activation function, residual quadratic networks with at most four neurons in one layer is universal approximator. \\
\indent Proof: As Figure 3 shown, because lemma 4 makes sure that the quadratic ReLU network can approximate function $K(\frac{||x-z_i||}{\sigma}), i=1,2...$. We use residual connections (red lines) to make the network modules that represent different $K(\frac{||x-z_i||}{\sigma}), i=1,2...$ and we use another groups of residual connections (green lines) to aggregate the outputs of these modules, such that quadratic ReLU network can represent any function from the family $S_K$ at any accuracy. By triangle relationship, we conclude that residual quadratic ReLU network with no more than four neurons in each layer is universal approximator. \\

\begin{figure}
  \centering
  \includegraphics[scale=0.27]{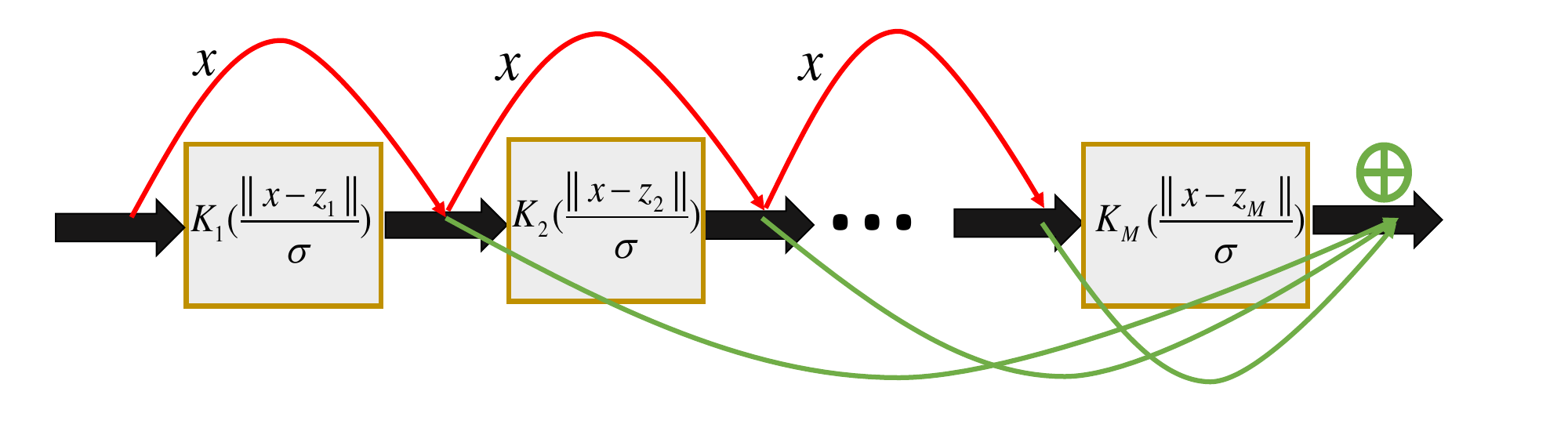}  
  \caption{Quadratic ReLU network with no more than four neurons using shortcuts in each layer is universal approximator. The red shortcuts are used for passing the input and the green shortcuts for aggregating the outputs of different modules.}
\end{figure}

\indent Compared with the results in [16] and [22], our Corollary 2 presents another special width-bounded universal approximator, which is slimmer than that of [16] and has sparser shortcuts than that of [22]. The upper limit of the width required by our width-bounded universal approximator is $4$, while the counterpart in [16] is $n+4$ given an $n-$input. Although only one neuron is demanded in each layer in [22], much denser residual shortcuts are employed as the trade-off. In our work, only sparser residual connections are needed to integrate different blocks. \\

\textbf{\large{3.3 Theorem 3}}\\
\indent \textbf{Key Idea for Proving Theorem 3:}  For universal approximation by neural networks, the current mainstream strategies are all based on piecewise approximation in terms of $L_1$, $L_\infty$, or other distance measures. For the purpose of piecewise approximation, the functional space is divided into numerous hypercubes, which are intervals in the one-dimensional case, according to a specified accuracy to approximate a target function in every hypercube. With quadratic neurons and the trick that two ReLU neurons can organically execute a linear operation, we can instead use a global approximating method that is much more efficient. At the same time, the quadratic network structure is neither too wide nor too deep, which can be regarded as a size-bounded universal approximator, in contrast to what we have introduced before. What's more, aided by Algebraic Fundamental theorem, our novel proof reveals the uniqueness and facility of our proposed quadratic networks that cannot be elegantly made by networks with quadratic activation. More importantly, Theorem 3 gives a first-of-the-kind global universal approximator, facilitating feature representations in deep learning. \\

 First, we show any univariate polynomial of degree N can be exactly expressed by a quadratic network with a complexity of $\mathcal{O}(log_2(N))$ in depth. Next we refer the result [34] regarding Hilbert's thirteen problem that multivariate functions can be represented with a group of separable functions, and then finalize the proof.\\

\indent \textbf{Lemma 5:} With ReLU as activation function, any univariate polynomial of degree $N$ can be perfectly computed by a quadratic network with depth of $\mathcal{O}(log_2(N))$ and width of no more than $N$.\\
\indent Proof: According to the Algebraic Fundamental Theorem [35], a general univariate polynomial of degree $N$ can be expressed as $P_{N}(x)=C\prod_{i}^{l_1}(x-x_i)\prod_{j}^{l_2}(x^2+a_{j}x+b_{j})$, where $l_{1}+2l_2=N$. The ReLU activation has an important property: $f(x)=ReLU(f(x))-ReLU(-f(x))$, which is often resorted in the literature, such as [48]. The network we construct is shown in Fig. 4 (Each neuron actually represents two parallel neurons to implement a linear operation). Every two neurons are grouped in the first layer to compute $(x-x_i)$, $(x-x_i)(x-x_{i+1})$ or $x^2+a_{i}x+b_i$, then the second layer merely uses  half the number of neurons in the first layer to combine the outputs of the first layer. By repeating such a process, the quadratic network with the depth of $O(log_2(N))$ can exactly represent $P_{N}(x)$. \\

\indent The following lemma shows that any univariate function $f(x)$ continuous in $[0,1]$ can be approximated with a Bernstein polynomial up to any accuracy, which is the Bernstein version of the proof for the Weierstrass theorem. \\

\indent  \textbf{Lemma 6:} Let $f(x)$ is a continuous function over $[0,1]$, we have
\begin{equation*}
\lim_{n \to +\infty} B_{n}(x)=f(x)
\end{equation*}
\indent Proof: It is well known; please refer to [33].\\

\indent \textbf{Corollary 2:} Any continuous univariate function supported on $[0,1]$ can be approximated by a quadratic network with ReLUs up to any accuracy.\\

\begin{figure}
  \centering
  \includegraphics[scale=0.27]{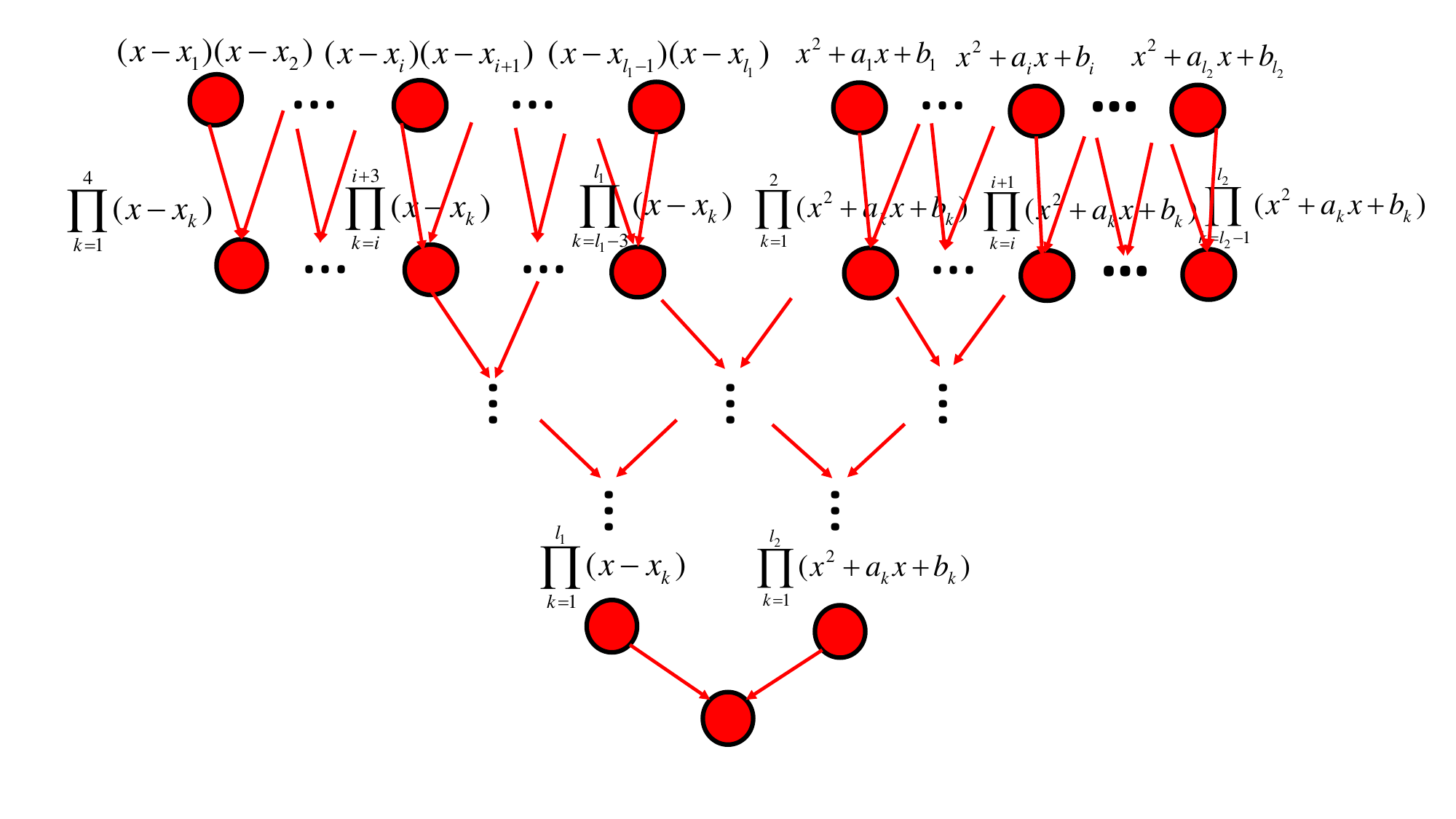} 
  \caption{Quadratic network approximating a univariate polynomial according to the Algebraic Fundamental Theorem.}
\end{figure}

\indent \textbf{Lemma 7:} Every continuous $n$-variable function on $[0,1]^n$ can be represented in the form: 
\begin{equation*}
f(x_1,...,x_n)=\sum_{i=1}^{2n+1}g_i(\sum_{j=1}^{n}h_{ij}(x_j))
\end{equation*}
\indent Proof: This is a classical theorem made by Kolmogorov and his student Arnold. Please refer to [34].\\

\indent \textbf{Proof of Theorem 3:} Combining Lemmas 5-7, it can be shown that the ReLU-embedded quadratic network is a universal approximator, and importantly it realizes a universal approximation in a global manner. \\

Let us look at the following numerical example to illustrate our novel quadratic network based factorization method. First, 100 instances were sampled of a function, $g(x)=(x^2+1)(x-1)(x^2+1.7x+1.2)$ in [-1,0]. Instead of taking opposite weights and biases to create linearity as used in proof for clarity, here we have incorporated shortcuts to make our factorization method trainable in terms of adaptive offsets. Using the ReLU activation function, we trained a four-layer network 4-3-1-1 with shortcuts to factorize this polynomial, as shown in Fig. 5. The parameters in connections marked by green symbols are fixed to perform multiplication. The shortcut connections and vanilla connection are denoted by green and red lines respectively. The neurons in the first layer will learn the shifted factors $T_1,T_2,T_3$, with unknown  constant offsets $C_1,C_2,C_3$. The whole network will be combined in pairs to form $T_1T_2,T_2 T_3,T_1 T_3$ in the next layer.  Then, the neurons in the third layer will multiply $T_1$ and $T_2 T_3$ to obtain $T_1T_2T_3$. Finally, the neuron in the output layer is a linear one that will be trained to undo the effect of the constant offsets aided by the shortcuts. We trained the network to learn the factorization by initializing the parameters multiple times, with the number of iterations 600 and the learning rate 2.0e-3. The final average error is less than 0.0051.  In this way, the function was learned to be $g^{'} (x)=(1.0117592x^2+0.02033176x+1.0040002)(0.00819827x^2+0.9977611x-1.0023365)(1.0113174x^2+1.7042247x+1.1885077)$, which agrees well with the target function $g(x)=(x^2+2)(x-1)(x^2+1.7x+1.2)$.  \\

Mathematically, we can handle the general factorization representation problem as follows. Let us denote $A_i=m_i, A_{ij}=m_im_j, A_{ijk}=m_im_jm_k,...,A_{123...n}=\prod_i^n m_i$, which are generic factors for $i=1,2,...,n$, the question becomes if $\prod_i^n (m_i+C_i)-\prod_i^n m_i$ can always be represented as the linear combination of  $A_0, A_i, A_{ij}, A_{ijk},...,A_{123...(n-1)}$ and a constant bias? If the answer is positive, then our above-illustrated factorization method can be extended for factorization of any univariate polynomial. Here we offer a proof by mathematical induction.\\

For $n=1$, we have $m_1+C_1-m_1=C_1$. Assume that we can represent $\prod_i^p (m_i+C_i)-\prod_i^p m_i$ with a linear combination of  $A_i, A_{ij}, A_{ijk},...,A_{123...(p-1)}$ and a constant, denoted as $f(A_i, A_{ij}, A_{ijk},...,A_{123...(p-1)})$. Then, $\prod_i^{p+1} (m_i+C_i)-\prod_i^{p+1} m_i=(m_{p+1}+C_{p+1})(\prod_i^p (m_i+C_i)-\prod_i^p m_i)+C_{p+1}\prod_i^p m_p=(m_{p+1}+C_{p+1})f( A_i, A_{ij}, A_{ijk},...,A_{123...(p-1)})+C_{p+1}\prod_i^p m_p$. Clearly, the above terms $m_{p+1}f(A_i, A_{ij}, A_{ijk},...\\,A_{123...(p-1)})$, $C_{p+1}\prod_i^p m_p$ are both linear terms of $A_i, A_{ij}, \\A_{ijk},...,A_{123...p}$ and  $C_{p+1}f(A_i, A_{ij}, A_{ijk},...,A_{123...(p-1)})$ is a linear representation of $A_i, A_{ij}, A_{ijk},...,A_{123...(p-1)}$ plus a constant bias as well. Combining these together, we immediately have a desirable linear representation of  $\prod_i^{p+1} (m_i+C_i)-\prod_i^{p+1} m_i$ that concludes our proof.\\

\begin{figure}
  \centering
  \includegraphics[scale=0.4]{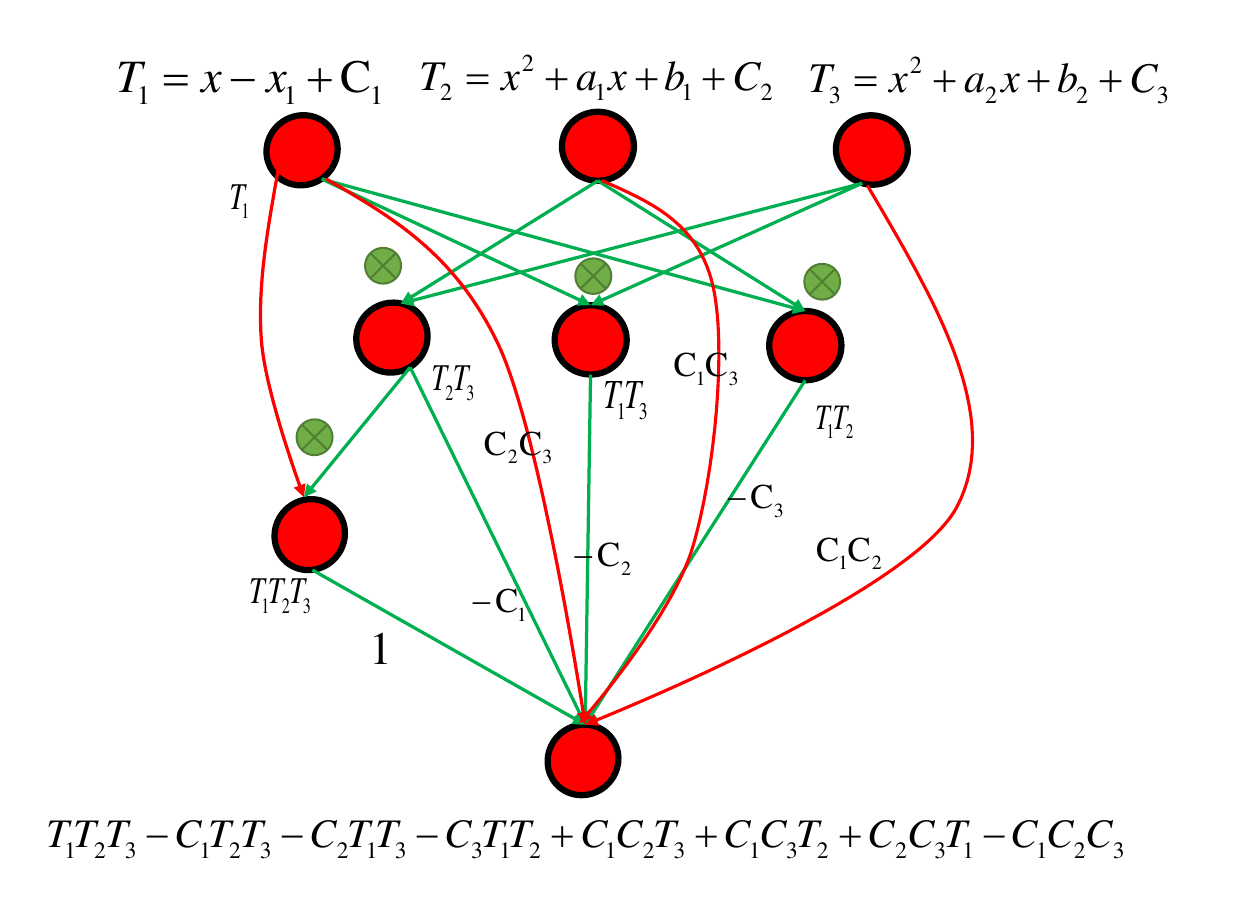}  
  \caption{Quadratic network with shortcuts to learn the soft factorization of an exemplary univariate polynomial.}
\end{figure}

\textbf{\large{3.4 Theorem 4}}

\indent \textbf{Key Idea for Proving Theorem 4: } It is notable that a quadratic neuron internally realizes addition and multiplication operations in a unique way, which enables quantized quadratic neurons as building blocks for construction of an arbitrary function in a top-down way. That is the key mechanism  by which the saving of $\mathcal{O}\left(\lambda\left(log^{\frac{1}{\lambda-1}+1}(\frac{1}{\epsilon})\right)(\frac{1}{\epsilon})^{\frac{d}{n}}\right)$, $\lambda \geq 2$ is achieved in terms of the number of weights. Figure 6 clearly demonstrates our heuristics.\\

\indent \textbf{Lemma 8:} A connection with a weight $w \in [-1,1]$ can be approximated by a ReLU network with $\lambda \geq 2$ distinct weights, satisfying that (i) the ReLU network is equivalent to weight $w'$ with error no more than $2^{-t}$, specifically, $|w-w'|<2^{-t}$; (ii) the depth is $\mathcal{O}[log(t)]$; (iii) the width is $\mathcal{O}(t)$; and (iv) the number of weights is at most $\mathcal{O}(t)$. \\
\indent Proof: We first consider $\lambda=2$ (we select two different weights as $2^{-1}$, $-2^{-1}$), $w \geq 0$ and $x \geq 0$, and will relax the constraints later. Setting $w=0$ directly leads to an empty operation. There are three other necessary operations perfectly represented by a ReLU quadratic network: $E_u$ (unity operation) $E_p$ (power operation) and $E_m$ (multiplication), which take 4, 8, 8 neurons respectively, as shown in Figure 6. Our goal is to approximate $w$ in accuracy of $2^{-t}$. Let us define 
\begin{equation*}
K_c \equiv\{2^{-1},2^{-2},...,2^{-(t-1)},2^{-t}\}.
\end{equation*}
With a homogeneous weight $2^{-1}$, all the members in $K_c$ can be represented by the composition of $E_u$, $E_p$ and $E_m$ operations. Because a binary expansion of $W_c$ can denote any integral multiple of $2^{-t}$ like a numeral system with the radix of $2^{-t}$, which is equivalent to represent any number with an error bound of $2^{-t}$.  To implement a binary expansion, one more layer with unity and empty operations is needed. Therefore, the overall quantized quadratic network has $log(t)+1$ layers, with no more than $24t$ weights. \\

\indent Regarding the situation of $w<0$, we are able to construct $-1$ with $-2^{-1}$ as the final layer. To relax the assumption $x \geq 0$, we can use two parallel sub-networks that take $x$ and $-x$ respectively. Due to the gate property of ReLU, the output sign of a sub-network can be flipped in the last layer. In the case of $\lambda > 2$, because our construction is top-down, it is interesting that $\frac{1}{2}$ and $-\frac{1}{2}$ are enough for the construction. We can assign two weight are $\frac{1}{2}$ and $-\frac{1}{2}$ respectively among $\lambda$ distinct weights. Therefore $\lambda$ does not affect the final complexity. \\

\begin{figure}
  \centering
  \includegraphics[scale=0.3]{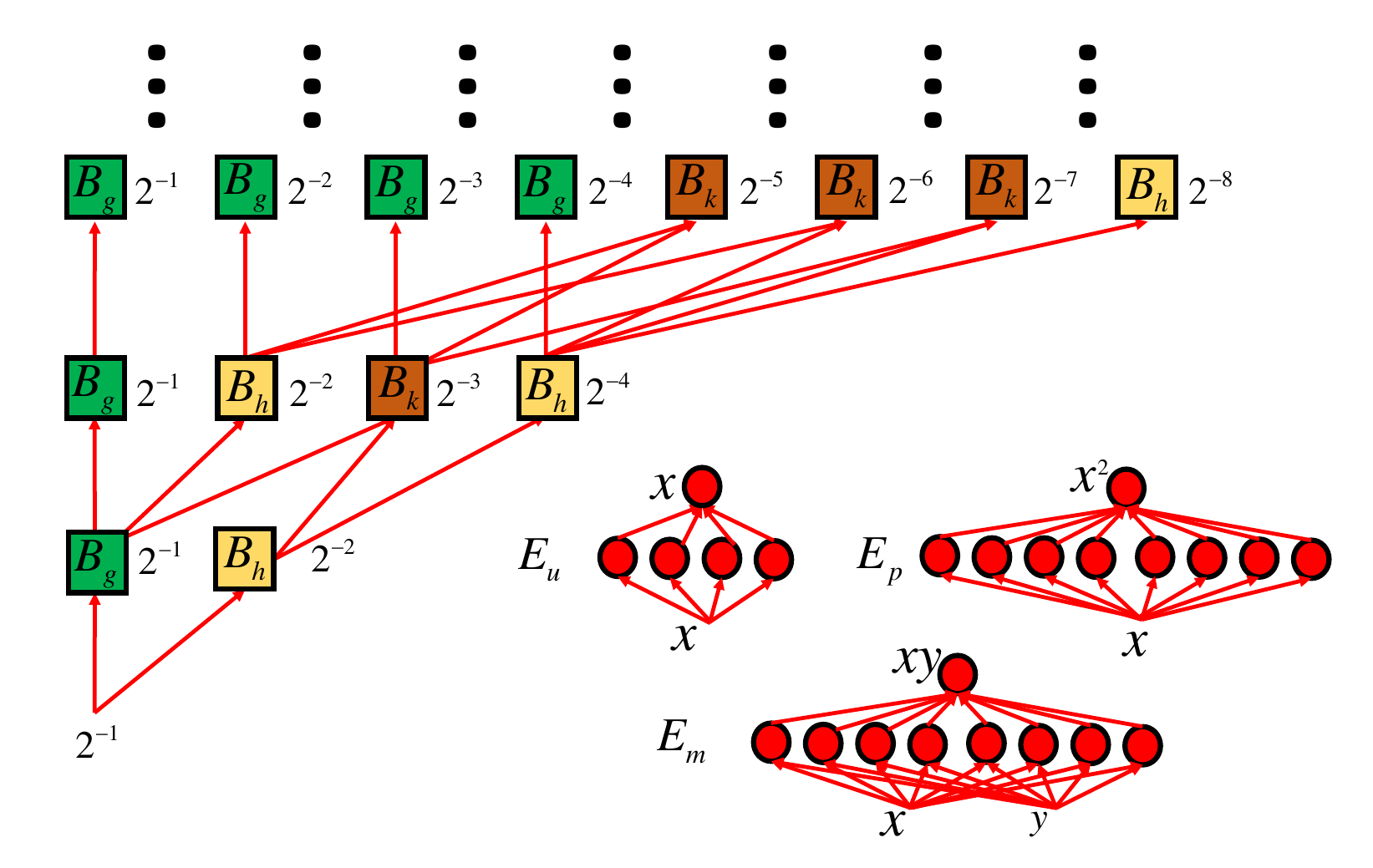}  
  \caption{Quantized quadratic ReLU networks with weight $2^{-1}$ and $-2^{-1}$ to approximate any number with an error bound of $2^{-t}$}
\end{figure}

\indent \textbf{Proof of Theorem 4:} The proof consists of the following two steps. First, we utilize $f'$ to approximate $f \in \mathcal{F}_{d}^{n}$, and then we construct $f'$ precisely with our quantized ReLU quadratic networks. For $\textbf{m}=(m_1,...,m_d) \in \{0,1,...,N\}^d$, $\psi_{\textbf{m}}(\textbf{x})$ is defined as:
\begin{equation*}
\psi_{\textbf{m}}(\textbf{x})=\Pi_{k=1}^{d} h(3Nx_k-3m_k),
\end{equation*}
where $N$ is a constant and 
\begin{equation*}
h(x) = \left \{
             \begin{array}{lr}
             1            \ \        |x| \leq 1 \\
             2-|x|        \ \     1 \leq |x| \leq 2  \\  
             0            \ \ |x|\geq 2
             \end{array}
\right.
\end{equation*}
$\psi_{\textbf{m}}(\textbf{x})$ forms a partition of the unity on $[0,1]^d$: $\sum_{\textbf{m}}\psi_{\textbf{m}}(\textbf{x}) \equiv 1,$ where $\textbf{x}$ is supported on $[0,1]^d$. Note that the support of $\psi_{\textbf{m}}(\textbf{x})$ is $\{\textbf{x}:|x_k-\frac{m_k}{N}| < \frac{1}{N}, \forall k\}$. For any $\textbf{m}$, there is a corresponding order $n-1$ Taylor expansion of the function $f$ at $x=\frac{\textbf{m}}{N}$: 
\begin{equation*}
Q_{\textbf{m}}(\textbf{x}) = \sum_{\textbf{n}:|\textbf{n}|<n} \frac{D^{n}f}{\textbf{n}!}|_{\textbf{x}=\frac{\textbf{m}}{N}}(x-\frac{\textbf{m}}{N})^{\textbf{n}},
\end{equation*}
where $\textbf{n}!=\Pi_{k=1}^{d} n_{k}!$, and $(x-\frac{\textbf{m}}{N})^{\textbf{n}} =\Pi_{k=0}^{d}(x_k-\frac{m_k}{N})^{n_k}$. 
Furthermore, we define $Q_{\textbf{m}}^{'}(\textbf{x}) = \sum_{\textbf{n}:|\textbf{n}|<n} \gamma_{\textbf{m},\textbf{n}}(x-\frac{\textbf{m}}{N})^{\textbf{n}}$, where $\gamma_{\textbf{m},\textbf{n}}$ denotes some integral multiple of $\frac{1}{n}(\frac{d}{N})^{n-|\textbf{n}|}$ that is the closest to $\frac{D^{n}f}{\textbf{n}!}|_{\textbf{x}=\frac{\textbf{m}}{N}}$. Thus, the approximation of $f$ is constructed as: 
\begin{equation*}
f^{'}(\textbf{x}) = \sum_{\textbf{m}}\psi_{\textbf{m}}Q_{\textbf{m}}^{'}(\textbf{x})
\end{equation*}
Hence, we have
\begin{equation*}
\begin{split}
&\sup_{\textbf{x} \in [0,1]^d } |f(\textbf{x})-f^{'}(\textbf{x})|  \\
&= \sup_{\textbf{x} \in [0,1]^d }|\sum_{\textbf{m}}\psi_{\textbf{m}}(f(\textbf{x})-Q_{\textbf{m}}^{'}(\textbf{x}))| \\
& \leq  \sum_{\textbf{m}} \sup_{\textbf{x} \in \{|x_k-\frac{m_k}{N}| < \frac{1}{N}, \forall k \} } |f(\textbf{x})-Q_{\textbf{m}}^{'}(\textbf{x})|  \\
& \leq 2^{d} \max_{\textbf{x} \in \{|x_k-\frac{m_k}{N}|< \frac{1}{N}, \forall k\}} |f(\textbf{x})-Q_{\textbf{m}}(\textbf{x})| \\
\ \ \ \ \ \ \ \ \ \ \ & +2^{d}\max_{\textbf{x} \in \{|x_k-\frac{m_k}{N}|< \frac{1}{N}, \forall k \}}|Q_{\textbf{m}}^{'}(\textbf{x})-Q_{\textbf{m}}(\textbf{x})| \\
\end{split}
\end{equation*}

\begin{equation*}
\begin{split}
& \leq \frac{2^{d}d^{n}}{n!}(\frac{1}{N})^n \max_{\textbf{n}:|\textbf{n}|=n} {ess\sup}_{\textbf{x} \in [0,1]^d} |D^{\textbf{n}}f(\textbf{x})| \\
& +2^d \sum_{\textbf{n}:|\textbf{n}|<n} \max{\bigg(|\frac{D^{n}f}{\textbf{n}!}-\gamma_{\textbf{m},\textbf{n}}|\bigg)} (x-\frac{\textbf{m}}{N})^{\textbf{n}}\\
& \leq \frac{2^{d}d^{n}}{n!}(\frac{1}{N})^n (\frac{1}{N})^n + \frac{2^{d}}{n}\bigg((\frac{d}{N})^n+...+(\frac{d}{N})^1(\frac{d}{N})^{n-1}\bigg)\\
& \leq 2^d (\frac{d}{N})^{n}(1+\frac{1}{n!})  \leq 2^{d+1} (\frac{d}{N})^{n} \\
\end{split}
\end{equation*}

Therefore, if we choose $N \geq (\frac{2^{d+1}d^{n}}{\epsilon})^{\frac{1}{n}}$, for $\epsilon \in (0,1)$, any $f\in \mathcal{F}_{d,n}$ can be approximated by $f^{'}$ with an error bound $\epsilon$. We rewrite 
\begin{equation*}
\begin{split}
f^{'}(\textbf{x}) & = \sum_{\textbf{m}\in \{0,1,...,N\}^d} \sum_{\textbf{n}: |\textbf{n}|<n} \gamma_{\textbf{m},\textbf{n}} \psi_{\textbf{m}} (x-\frac{\textbf{m}}{N})^{\textbf{n}} \\
&  = \sum_{\textbf{m}\in \{0,1,...,N\}^d} \sum_{\textbf{n}: |\textbf{n}|<n} 
\gamma_{\textbf{m},\textbf{n}} f_{\textbf{m},\textbf{n}}^{'} (\textbf{x}), \\
\end{split}
\end{equation*}
where  $f_{\textbf{m},\textbf{n}}^{'} (\textbf{x}) = \psi_{\textbf{m}} (x-\frac{\textbf{m}}{N})^{\textbf{n}}.$ $f^{'}(\textbf{x})$ is the linear combination of no more than $d^{n}(N+1)^d$ terms of $f_{\textbf{m},\textbf{n}}^{'}(\textbf{x})$. By Lemma 8, with $t=log (\frac{nN^{n}}{d^n})$ we can use a sub-network of $\mathcal{O}[log (t)]$ to represent the constant weights $\gamma_{\textbf{m},\textbf{n}}$. Then, the remaining work is to construct $f_{\textbf{m},\textbf{n}}^{'} (\textbf{x})$. $N$ is chosen as $2^{l}$, where $l$ is some large integer, such that all $\frac{m_i}{N}$ can be obtained precisely in the way of Lemma 8. \\

The strategy of using a quadratic network to construct $f^{'}$ is illustrated in Figure 7. Table 1 presents the complexities of individual blocks of interest, which helps us to analyze the complexity of the whole network. Although the whole complexity can be expressed in an explicit way, here we use the $\mathcal{O}$ notation for clarity and simplicity. Let $N_d$ and $N_w$ be the network depth and the number of weights. According to Table 1, we have $N_d = \mathcal{O}\left(log(t)+log(N)\right)$ and $N_w = \mathcal{O}\left(N^{d}log(t)+N^{d}log(N)\right)$. Because $t=log(\frac{nN^n}{d^n})$ and $N= (\frac{2^{d+1}d^{n}}{\epsilon})^{\frac{1}{n}}=\mathcal{O}\left((\frac{1}{\epsilon})^{\frac{1}{n}}\right)$, substituting them into the estimates results in
\begin{equation*}
\begin{split}
& N_d = \mathcal{O}\left(log(log(\frac{1}{\epsilon}))+log(\frac{1}{\epsilon})\right) \\
& N_w = \mathcal{O}\left(log(log(\frac{1}{\epsilon}))(\frac{1}{\epsilon})^{\frac{d}{n}}+ log(\frac{1}{\epsilon})(\frac{1}{\epsilon})^{\frac{d}{n}}\right)
\end{split}
\end{equation*}

\begin{figure}
  \centering
  \includegraphics[scale=0.45]{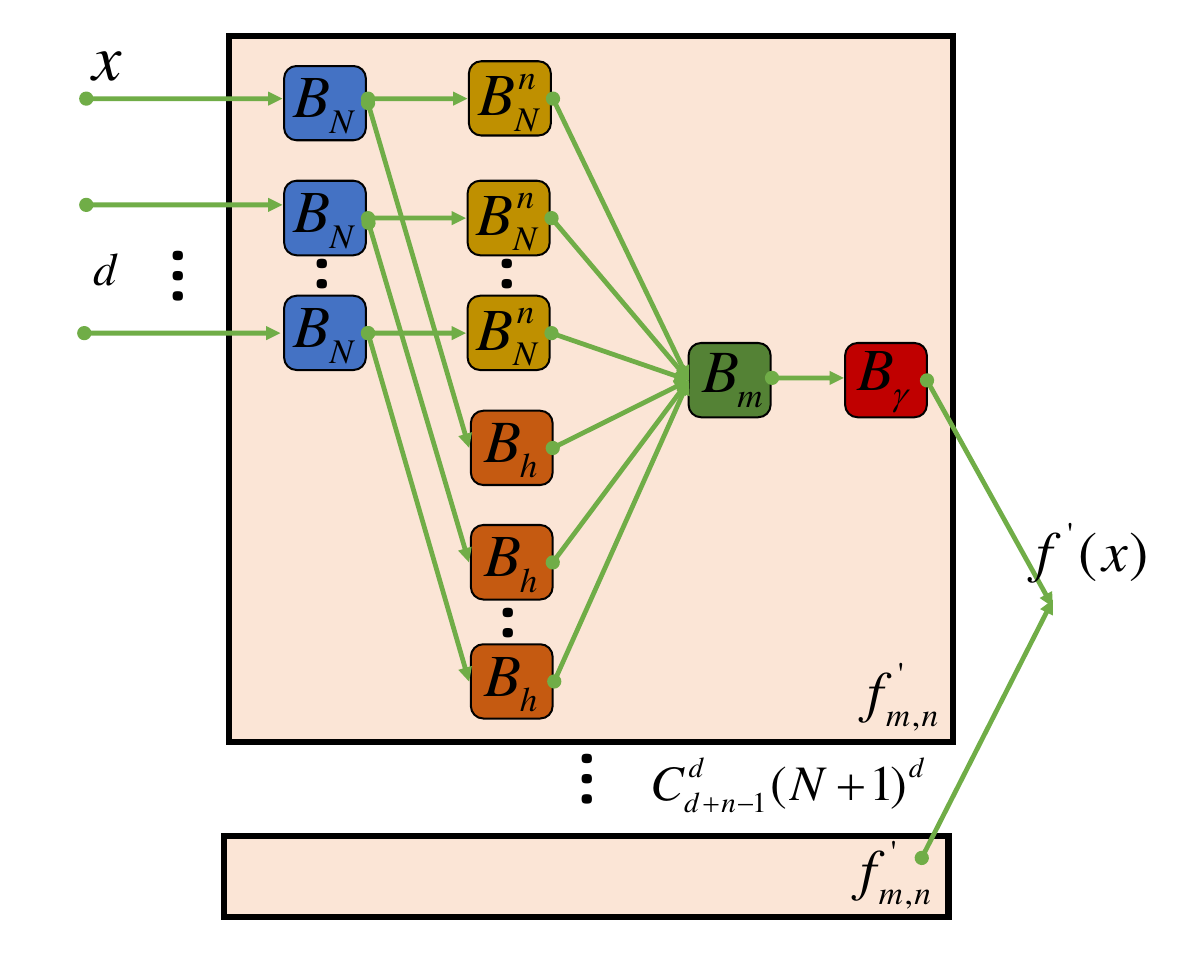}  
  \caption{Strategy of using a quantized quadratic network to implement $f^{'}(\textbf{x})$ }
\end{figure}

\begin{table}
\centering
{Table I: Descriptions of different building blocks.} \label{tab:title} 
 \begin{tabular}{||c c c c||} 
 \hline
 Block & Operation & Width & Depth \\ [0.5ex] 
 \hline\hline
 $B_N$ & Construct $\frac{1}{N},...,\frac{N-1}{N}$ & $N$ & $log(N)$ \\ 
 $B_N^n$ & Construct $(x_k-m_k)^{n_{k}}$ & $n$ & $log(n)$ \\
 $B_h$ & Implement $h(x)$ & 12 & 1\\
 $B_m$ & Multiply $d$ terms  & $2d$ & $log(2d)$ \\
 $B_{\gamma}$ & Construct $\gamma$ & $t$ & $log(t)$ \\ [1ex] 
 \hline
 \end{tabular}

\end{table}

 As we know, the quantization of the network [38] is an effective and robust way for model compression, even a binarized network performs well in some meaningful tasks. However, the reason why quantization is robust and useful was rarely studied. Ding et al. [39] reported that to obtain an approximation error $\epsilon$, the upper bound for a quantized ReLU network is delineated by $\mathcal{O}\left(\lambda\left(log^{\frac{1}{\lambda-1}+1}(\frac{1}{\epsilon})\right)(\frac{1}{\epsilon})^{\frac{d}{n}}+log(\frac{1}{\epsilon})(\frac{1}{\epsilon})^{\frac{d}{n}}\right)$, where $\lambda$ is the number of distinct weights. Because it is convenient for quadratic neurons to implement multiplication, our quantized ReLU quadratic network considerably reduces the upper bound to $\mathcal{O}\left(log\left(log(\frac{1}{\epsilon})\right)(\frac{1}{\epsilon})^{\frac{d}{n}}+ log(\frac{1}{\epsilon})(\frac{1}{\epsilon})^{\frac{d}{n}}\right)$. It is observed that $\mathcal{O}\left(log\left(log(\frac{1}{\epsilon})\right)(\frac{1}{\epsilon})^{\frac{d}{n}}\right)$ is significantly lower than $\mathcal{O}\left(\lambda\left(log^{\frac{1}{\lambda-1}+1}(\frac{1}{\epsilon})\right)(\frac{1}{\epsilon})^{\frac{d}{n}}\right)$, therefore the saving made by quantized quadratic network is almost in an order of $\mathcal{O}\left(\lambda\left(log^{\frac{1}{\lambda-1}+1}(\frac{1}{\epsilon})\right)(\frac{1}{\epsilon})^{\frac{d}{n}}\right)$ . More interesting, the complexity bound achieved by quantized quadratic networks is independent of the value of $\lambda$, as explained in our proof. \\

\section{Discussions and Conclusion}

\indent The rational for the first theorem is easily understandable. Since it is commonly known that a conventional network with one hidden layer is a universal approximator, the best thing we can do to justify the quadratic network is to find a class of functions that can be more efficiently approximated by a quadratic network than with a conventional network. By the first theorem, the quadratic network is indeed more efficient than the conventional counterpart.\\

\indent The previous studies demonstrate that any function of $n-$variables that is not constant along any direction cannot be well represented by a fully-connected ReLU network with no more than $n-1$ neurons in each layer [16]. However, breaking this network width bound, our second theorem states that when a radial function is not constant in any direction, the network width $4$ is sufficient for a ReLU quadratic network to approximate the function accurately. Since it is more convenient to represent a nonlinear function with more nonlinear neurons, the incapability of conventional networks can be compensated by quadratic networks. Theorem 2 implies a slimmer and/or sparser width-bounded universal approximator.  \\

\indent Our third theorem is inspiring that a general polynomial function can be exactly expressed by a quadratic network via ReLU activation in a novel way of data-driven network-based algebraic factorization. Although third theorem assumes the ReLU activation function, considering ReLU is arguably the most important activation function, it makes a great sense in illustrating the capacity of quadratic networks. Most importantly, Theorem 3 shows that quadratic networks are advantageous in terms of expressibility and efficiency. By calculating a quadratic function of input variables, quadratic neurons create second-order terms. The second-order terms serve as nonlinear building blocks that are different from non-linearity generated only through  nonlinear activation, and clearly preferred from the perspective of the Algebraic Fundamental Theorem. Accurately forming a polynomial allows global computation and high efficiency as compared to the popular piece-wise approximation with conventional neurons in a deep network. In our quadratic network construction, both the depth and width are limited, while all previous universal approximators based on ReLU are either too wide or too deep to achieve a high approximation accuracy. This fact suggests that quadratic networks could potentially minimize the number of neurons and free parameters for superior machine learning performance. It is surprising that although every quadratic neuron possesses a number of parameters three times as many as that of the corresponding conventional neuron, quadratic networks can use less parameters than the conventional counterparts in important cases, such as the presentation of radial functions and the approximation to multivariate polynomials.\\

\indent Interestingly, our results (Corollary 1 and Theorem 3) imply that quadratic networks are highly adaptive. For example, while a quadratic network can do piece-wise approximation by degenerating into conventional networks, a generic quadratic network is capable of approximating any function in a global way by implementing basic algebraic factors. Also, quadratic networks can mimick RBF networks on the "wavelet" scale, as indicated by Corollary 1.\\

\indent As we know, deep learning generalizes well in many applications. Some researchers questioned that quadratic networks may tend to over-fitting because the added neural complexity may not be necessary. With our above proved theorems, we refute that since quadratic networks \textit{de facto} are more compact and more effective than the conventional counterpart in major circumstances, quadratic networks could improve the generalization ability, according to Occam's razor principle. Since general features are either quadratic or can be expressed as  combinations or compositions of quadratic features, quadratic deep networks will be universal, instead of being restricted  to the circumstances where only quadratic features are relevant.\\

\indent In conclusion, we have analyzed the approximation ability of the quadratic networks and proved four theorems, suggesting a great potential of quadratic networks in terms of expressive efficiency, unique representation, compact structure and computational capacity. Future research will be focused on evaluating the performance of deep quadratic networks in typical applications.

\vskip100pt

\bio{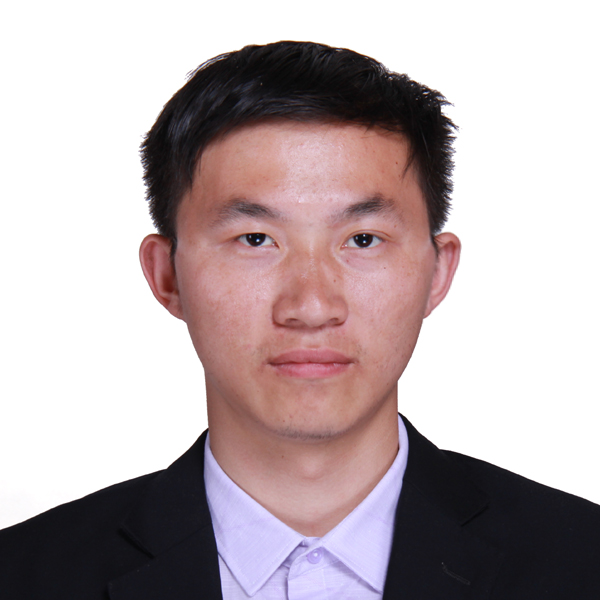}
Fenglei Fan received the Bachelor's degree from the Harbin Institute of Technology before joining Dr. G. Wang$'$s Laboratory. He is currently pursuing the Ph.D. degree with the Department of Biomedical Engineering, Rensselaer Polytechnic Institute. His research interests include applied mathematics, machine learning, and medical imaging.
\endbio

\bio{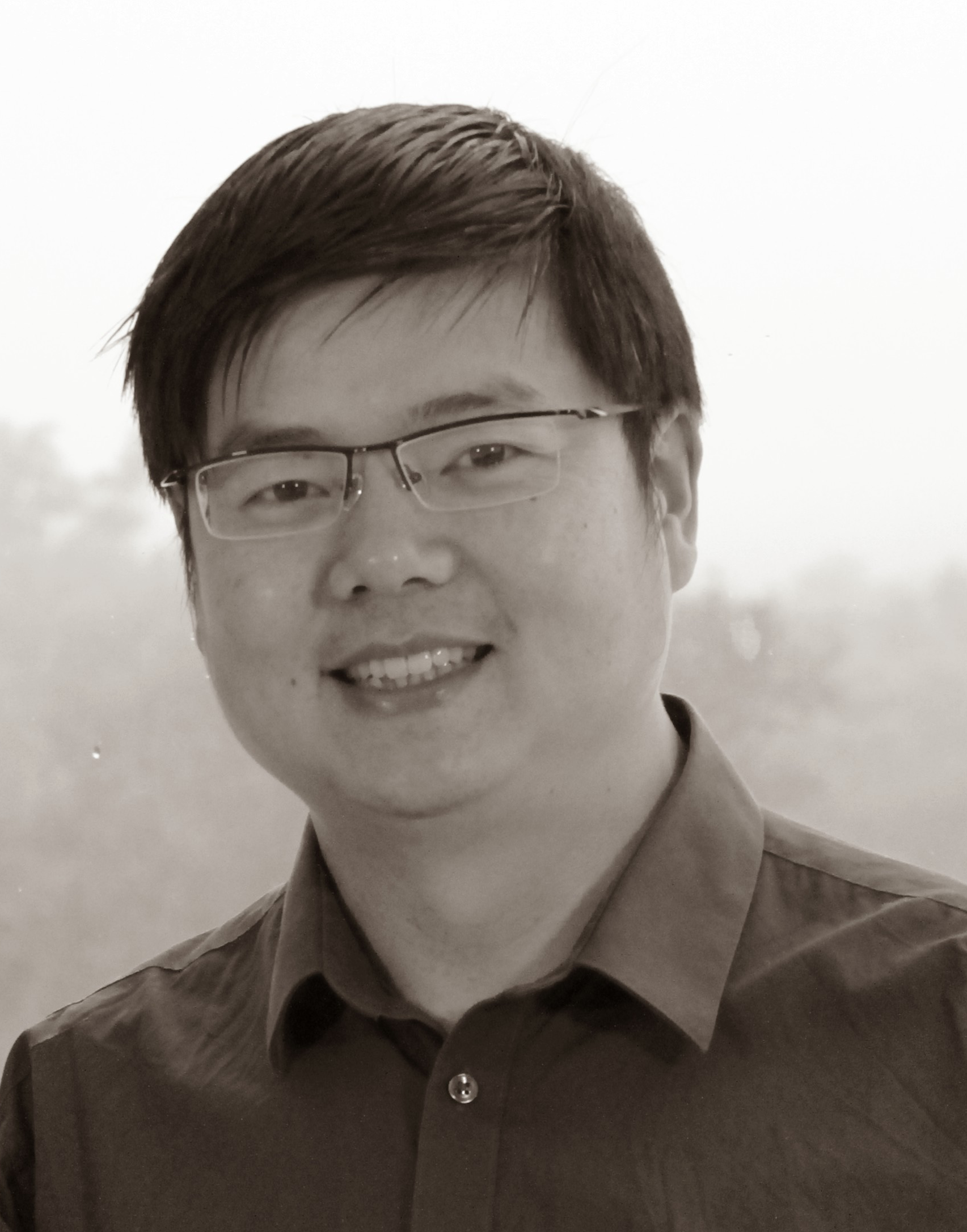}
Dr. Jinjun Xiong is a program director for cognitive computing systems research at the IBM T.J. Watson Research Center and a co-director of the IBM-Illinois Center for Cognitive Computing Systems Research. His research interests include cognitive computing, big data analytics, deep learning, smarter energy, and VLSI circuits and systems. Dr. Xiong received a PhD in electrical engineering from the University of California, Los Angeles. He has received four Best Paper Awards, one Best Paper in Track Award, and eight nominations for Best Paper Awards. He is an IBM Master Inventor and has received various IBM technical achievement awards and the IEEE Region One Outstanding Technical Contribution Award.
\endbio

\bio{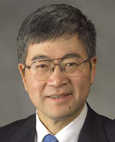}
Dr. Ge Wang is currently a Clark and
Crossan Chair Professor and the Director of the
Biomedical Imaging Center, Rensselaer Polytechnic
Institute, USA. He authored the papers
on the first spiral/helical cone-beam/multi-slice
CT algorithm. Currently, there are over 100 million
medical CT scans yearly with a majority in the
spiral cone-beam mode. He pioneered bioluminescence
tomography. His group published the
first papers on interior tomography and omnitomography
(all-in-one) to acquire diverse data sets simultaneously (all-at-once).
His results were featured in Nature, Science, and PNAS, and
recognized with awards. He has written more than 430 peer-reviewed journal
publications. He is a Fellow of SPIE, OSA, AIMBE, AAPM, and AAAS.
\endbio

\makeatletter

\def\pct{\expandafter\@gobble\string\%}

\immediate\write\@auxout{\pct\space This is a test line.\pct }

\end{document}